\documentclass[conference]{IEEEtran}  

\makeatletter
\let\NAT@parse\undefined
\makeatother

\IEEEoverridecommandlockouts                              

\usepackage{times}  
\usepackage{helvet}  
\usepackage{courier}  
\usepackage{url}  
\usepackage{graphicx}  
\usepackage{balance}
\usepackage{color}
\usepackage{amsmath}
\usepackage{amssymb}
\usepackage{multirow}
\usepackage[font={small}]{caption}
\usepackage[utf8]{inputenc}
\usepackage{booktabs}
\usepackage{subfig}
\usepackage{dblfloatfix}
\usepackage{tikz}
\usepackage{cite}
\usepackage{algorithm}
\usepackage{algpseudocode}
\usepackage{cancel}
\usepackage{todonotes}
\usepackage{xspace}
\usepackage{microtype}
\usetikzlibrary{shapes, arrows}
\usetikzlibrary{backgrounds, fit, calc, 3d}
\usetikzlibrary{positioning}
\definecolor{dred}{rgb}{1,0,0}
\definecolor{dblue}{rgb}{0,0,1}

\renewcommand{\baselinestretch}{0.99}

\newcommand{\ie}{\mbox{i.\,e.}\xspace}

\renewcommand{\eqref}[1]{Eq.~(\ref{#1})}
\newcommand{\figref}[1]{Fig.~\ref{#1}}

\title{Doing Right by Not Doing Wrong in\\Human-Robot Collaboration
\thanks{$^\ast$These authors contributed equally. All  authors  are  with the Department of Computer Science, University of Freiburg, Germany. This work has been supported by the BrainLinks-BrainTools center of the University of Freiburg.}
}

\author{\IEEEauthorblockN{Laura Londoño$^\ast$}
\and
\IEEEauthorblockN{Adrian Röfer$^\ast$}
\and
\IEEEauthorblockN{Tim Welschehold}
\and
\IEEEauthorblockN{Abhinav Valada}
}

\begin{document}

\maketitle

\begin{abstract}
As robotic systems become more and more capable of assisting humans in their everyday lives, we must consider the opportunities for these artificial agents to make their human collaborators feel unsafe or to treat them unfairly.
Robots can exhibit antisocial behavior causing physical harm to people or reproduce unfair behavior replicating and even amplifying historical and societal biases which are detrimental to humans they
interact with. In this paper, we discuss these issues considering sociable robotic manipulation and fair robotic decision making. We propose a novel approach to learning fair and sociable behavior, not by reproducing positive behavior, but rather by avoiding negative behavior. In this study, we highlight the importance of incorporating sociability in robot manipulation, as well as the need to consider fairness in human-robot interactions.
\end{abstract}


\section{Introduction}
We envision a future in which robots can be ubiquitous helpers in human environments and assist humans with everyday tasks. Aside from the many challenges in robotic perception, planning, and manipulation that need to be addressed, we also see a need to address the societal impact that these robots will have. In recent years, the discussion in the AI community about AI perpetuating harmful stereotypes, and cementing existing systemic discrimination has grown louder~\cite{nelson2019bias,yu2019framing}. 
Existing AI systems have been shown to be lacking in these regards and the field focused on fairness-aware learning has formed, that seeks to analyze AI's potential for exhibiting discriminatory behavior and to propose means to remedy it.

In the field of robotics, these efforts are still in their infancy, largely due to there not being many scenarios of human-robot-interaction yet in which the robot actually perceives the human and changes its actions based on its perceptions.
As humans and robots start to act in closer proximity, either working in parallel, or collaborating on tasks, the potential for robot induced harm grows. Setting aside potential physical injury, we must also be conscious of the fact that robot behavior might make people feel unsafe, disrespected, rejected, or humiliated. Arguably, behavior resulting in these feelings will have a strong negative impact on humans, as they are exhibited during a personal, physical interaction and ultimately even lead to a diminishing acceptance of robots in human environments.   

\begin{figure}
    \centering
    \includegraphics[width=8.5cm]{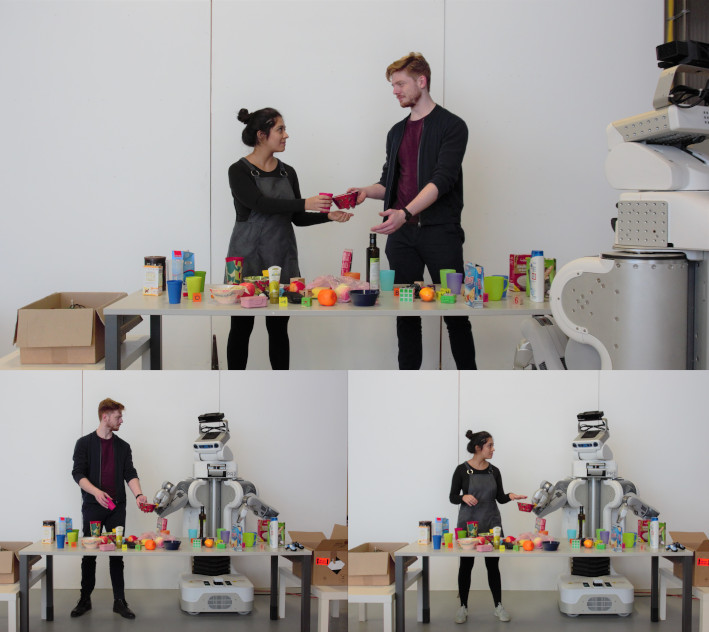}
    \caption{Robot observing two people collaborate on a sorting task and then having to interact suitably with both of them. The two people differ significantly both in their physical and their personal attributes. Is it fair for the robot to uncritically mimic the demonstration?}
    \label{fig:teaser}
\end{figure}

This concern necessitates the question what \emph{good} robot behavior is. Are the motions the robot is performing pleasant to the humans in its vicinity? Are the decisions a robot takes fair to a human collaborator?
In this paper, we argue that it might not be prudent to optimize for the ideal sociable motions, or the fairest decisions, as these might be hard to determine. Instead, we argue that it is sufficient for robotic agents to learn what antisocial, or unfair behaviors are in order to limit their behavior to being social and fair.
We propose this argument in two areas of robotics: motion generation and decision making. Being embodied agents is a unique feature of robots and thus it must be considered carefully how executed motions are perceived by humans. 
In the area of decision making, we are concerned with the kinds of plans robots might generate and how these impact people in the short- and long-term. Here, we find parallels to general AI fairness research and examine utilizing protected characteristics positively, to facilitate equitable and inclusive robotic behavior.

For both of these areas, we ground the problem of social and fair robotic behavior in insights from sociological and psychological study. We highlight technical approaches either performing initial steps towards addressing these challenges, or approaches which can be employed to teach robots \emph{what not to do}, as per our proposed approach.

\section{Safe and Social Robot Manipulation}
Sociability is currently defined as the ability of people to interact peacefully and pleasantly with others based on the use of social norms and social skills. Importantly, in social environments there exist reciprocal influences between individuals and the social context~\cite{corradi2016determines}. On a high level, it is connected not only to friendliness but also with respectful behavior. Thus, researchers argue that social behavior transcends specific actions, and situation and distinguishes them based on concepts like social norms, and attitudes such as empathy, and respect~\cite{findlay2006links}. On the one hand, \emph{social norms} refer to not invading personal space, not blocking peoples' paths or actions, make direct eye contact with the person you are speaking with, do not interrupt someone while they are talking, among others. On the other hand, respectful behavior means the capacity to perform actions that do not endanger the dignity or the psychology of others. In spaces where small groups of people interact directly, such as family gatherings, outings with friends, as well as in large spaces where large groups of people interact, such as in parks, shopping malls, hospitals, bars, etc., people constantly and unconsciously make use of social norms, social skills and dignified behavior. The stability of the interaction of different social groups, as well as societies, depends on this social behavior. Consequently, social behavior plays an important role in maintaining social cohesion in smaller and larger groups of people, as well as in society itself. The representation of this social behavior depends on the postures, expressions and sometimes vocalisations which denote that an approach is friendly and not hostile~\cite{ewer1968amicable}. 


Roboticists have also investigated the sociability of robots for human-robot-interactions (HRI). One line of inquiries has investigated the imitation of social cues and appropriate communication in social contexts. Researchers have studied the impact of the physical appearance, reactions to touch, and communication routines on human participants~\cite{leite2013social,bicchi2015social, robert2018personality}. The focus for most of these efforts has been on deploying robots in health- or elder care settings. They have been successful in evoking an attribution of social intentions from study participants, and were able to demonstrate that robots can benefit social interactions. 
However, these works focus on social interaction as the primary task of the robot. This neglects the aspect of circumstantial social interaction. When we want to task a robot with helping a human to unpack groceries, fold laundry, or wipe the floor, these skills are commonly developed in isolation. Naturally, there have been inquiries into safe human-robot collaboration~\cite{lasota2017survey}. However, most of these are investigating physical safety which is the avoidance of physical harm to the human. As pointed out in~\cite{lasota2017survey}, \emph{psychological safety} is rarely investigated in robotic manipulation. Humans are very sensitive to robots obeying physical social conventions, to feel both safe and at ease around robotic agents. The authors of~\cite{lasota2017survey} summarize a few works that have studied these phenomena and point to others who have limited the solution space for the planned motions to a set of robot velocities, and accelerations which they empirically determined to be psychologically comfortable for humans.
The area in robotics, which has investigated integrating social constraints into its tasks most thoroughly, is likely the autonomous navigation community~\cite{charalampous2017recent} for whom respecting personal space, and abiding by human patterns of circumnavigation is a must.

As robotic manipulation skills improve, the question of acting sociably around humans becomes more and more prevalent. The field needs to develop an understanding of, and metrics for humans' perception of robotic motions. According to the metrics, our robotic agents must be able to learn or generate manipulation actions which remain responsive to the humans around the robot, and make them feel at ease, if not even comforted by the robot. While approaches for the prediction and avoidance of human motions exist~\cite{mainprice2013human} and have shown to greatly improve efficiency of human robot collaboration, a more general approach to generating socially compliant manipulation is needed for enabling general human robot cooperation and collaboration.

In robot navigation, researchers have derived an extensive set of constraints that specify the social aspect of traveling around people. Commonly, these constraints are based on social norms~\cite{kirby2010social,rios2015proxemics}. In~\cite{kirby2010social} the authors expose that traditional robot control algorithms for path planning and obstacle avoidance treat all unexpected sensor readings as subjects that must be avoided. These traditional avoidance methods do not abide by social norms. The authors present a list of relevant constraints for robot navigation around people such as minimizing distance, obstacle avoidance, people avoidance, personal space, robot space, passing on the right, default velocity, facing the direction of travel and inertia. These first three, minimizing distance and the two aspects of social avoidance, relate to the task of traveling to a goal. Personal space, robot space and passing to the right are connected with social aspects of navigating around people. Finally, default velocity, facing the direction of the travel and inertia are understood as both task and social conventions, because failing to observe them can be inefficient (task-related) and socially awkward (social conventions). 

While these constraints can be formulated rather easily in a 2D navigation task, the high-dimensional nature of robotic manipulation makes it difficult to derive such heuristics manually. Instead, robots will need to be able to learn bounds on their motions conditioned on the people around them. While detecting a human collaborator's comfort or discomfort in the presence of the robot might be too difficult for a first step, kinesthetic teaching, as done by~\cite{bajcsy2017learning}, might be a feasible manner to communicate discomfort to the robot. The authors of~\cite{chisari2021correct} demonstrate an online interactive scheme for updating a manipulation policy through human feedback. 
With the space of discomfort inducing motions and postures learned, the robot can restrict its motion to an agreeable set.

Enabling robot manipulation to exhibit these social skills can lead to smoother and more positive human-robot interaction~\cite{leite2013social,bicchi2015social}. In the absence, collaboration can be intrusive and awkward, and become uncomfortable. When robots behave in an antisocial manner this can cause the entire human-robot collaboration to break down. 


Once the robot is able to learn to restrict its range of motion, we will come to the realization that there will not be one space of agreeable motions for everyone. Instead, depending on a human partner's personality, and likely also protected characteristics, the robot will have to decide which limiting space is acceptable. When interacting with an elderly person, fast motions might be seen as impatient, while a younger person might perceive them as lazy. In the end, it should be our goal as roboticists to enable robots to accommodate the diverse preferences of the diverse human collaborators that might interact with it. We must avoid causing people discomfort due to a bias in the collected data, or a person's membership in a minority group with little statistical weight. In the following section, we discuss the aspect of  \emph{fair}, not equal, robotic behavior and decision making in the context of collaborative manipulation.

\section{(Un-) Fair Human-Robot Collaboration}
Fairness is presented as an ambiguous ethical concept since it has different interpretations depending on the field in which it is used. In machine learning and artificial intelligence, fairness is a established area that studies how to ensure that the biases in the data and model inaccuracies do not lead to AI that treat individuals unfavorably on the basis of protected characteristics such as race, gender, disability, socioeconomic and sociodemographic position~\cite{caton2020fairness}. Fairness has become an important subject of study in AI research, after it was discovered that many of the state of the art models display a tendency to perpetuate historical or systemic biases against disadvantaged groups~\cite{hamilton2019sexist,rudin2018age,hamilton2019biased}. To understand how these larger implications are reflected in small-scale human-robot interaction, consider a simple collaborative task that a robot could learn from demonstrations:

A human and a robot are tasked with tidying up a table together. As illustrated in \figref{fig:teaser}, both of them are going to stand on the same side of the table, while there are crates at the ends which the items scattered on the table need to be sorted into. In order to prepare for the task, the robot will observe human participant pairs perform it.
On the one hand, the robot will study the way the participants move in each other's presence, learning the acceptable range of motion, but it will also monitor the macroscopic actions taken by the participants: Which items do they pick up, where do they place them. In the presented setup with the crates at the far ends of the table, an efficient and fair solution would be for either participant to move the items belonging in the crate far away from them to the center of the table while placing all the items within their reach into the crate close to them.
In general, we would expect to see an equal amount of items moved by either partner. According to~\cite{bratman1992shared} a deviation from such behavior would be deemed unfair. While it will be difficult, maybe even impossible, to exhibit \emph{perfectly} fair behavior, we can teach the robot not to imitate demonstrations, which violate fairness grossly, implicitly teaching it to reproduce \emph{reasonably} fair behavior. However, there are cases in which this behavior might be perceived as unfair: If one of the collaborators has a physical disability, they might not be able to move an equal amount of items in the same time. This does not demonstrate an unfair treatment of their non-disabled partner. In turn, their collaboration partner might eventually enter the disabled partner's personal space, in order to take on some of the remaining workload. With an other collaboration partner, this overreach might be viewed as undue impatience. With a different set of protected characteristics, \ie mismatched gender, it might even be an expression of negative discrimination.

The design of fair algorithms in machine learning, as well as in robot learning, has become a crucial policy issue. Current legislation ensures fairness by barring algorithm designers from using demographic information in their decision making, pursuing an approach of \emph{fairness through unawareness}. Researchers have been critical of these measures~\cite{fu2021fair, johnson2021algorithmic}. They explain that developers employing equal treatment can lead to disparate impact given that there are important differences among groups of people based on demographic differences and protected characteristics.
As~\cite{hurtado2021learning} points out, the uncritical imitation of social interactions conditioned on protected characteristics does not prevent discrimination. Instead, they suggest a method to relearn a robotic behavior, which equalizes it with respect to selected protected characteristics.
However, a human supervisor is still needed to inform the approach of pairs of attributes with respect to which its policy is unfair.
Nonetheless, in machine learning, the use of impact parity in fair algorithms has been presented as a way to ensure real equal opportunities and treatment for diverse groups. As it is possible to observe in this discussion, the definition of fairness and use of protected characteristics still being a dilemma in machine learning and robot learning.
Firms, institutions, and governments are using machine learning algorithms in areas where people are expecting fair treatment. However, the difficulty of judging a fair decision has made this problem increasingly complex~\cite{corbett2018measure,caton2020fairness}. While it is possible in most of the institutional applications of AI to verify a deployed model's fairness statistically, in robot manipulation this will not be as easy. The range of exhibited motions and generated plans is highly dependent on the context in which the robot is operating. Further, there might be no statistical comparison once we enable our robots to learn new tasks demonstrated by individual users. Thus, we conclude that we will have to build on humans surrounding the robot telling it that it is behaving antisocially, or unfairly and for the robot to adjust it behavior and decisions instantly.

This inverse form of learning fair and sociable behavior is not as contrary as one might initially think. In daily life, people continuously judge actions as fair and unfair depending on the outcome of the action, moral or cultural norms, personal experience and bias. Generally, the standards of fairness used in social context are used in applied science such as robotics. For instance, in~\cite{chang2020defining} the theory of distributive procedural and psychology justice is used. However, even in social contexts people tend to emphasize these standards differently when accounting for the fairness-unfairness experiences versus those they have witnessed and when judging fair versus unfair outcomes~\cite{lupfer2000folk}. In the field of law, researchers consider it necessary to define unfairness in order to recognize fairness~\cite{herrine2021folklore}. In philosophy, philosophers such as David Hume and John Rawls state that unfairness is a basic element of ethical judgements in fields such as politics and morality~\cite{ryan2006fairness}. Therefore, the sense of fairness ought to yield an outcome that people would agree on. Consequently, fair decisions should abide by most rules against objectionable discrimination, \ie a person not being selected for a job based on aspects of their race or gender. Objectionable discrimination results in injustice, inequity, deceptive, and abusive behavior. Also, unfair behavior is commonly characterized to cause rejection in the society, and often yields a tendency to correct it.  Consequently one can argue that every action that impacts the society negatively, causes damage to other people, especially the most disadvantaged~\cite{finkel2001not}. In general, people tend to agree about the unfairness of presented situations and universally reject them. It is important to mention that a meaningful structural representation of unfairness has to consider the particular content of unfair situations as well as the social setting where they occur~\cite{mikula1990people} and the group, or individuals affected. 

Finally, from a technical perspective, problems involving multiple agents can be understood through the lens multi-agent reinforcement learning~\cite{zhang2021multi}. Recently, researchers have investigated the inclusion of \emph{social goals} as additional goals for agents~\cite{tejwani2022social}. While these goals are very primitive at the moment, they can produce supportive or antagonistic behaviors in agents. Going forward it might be possible to employ such models to infer the perception of fairness that a collaboration partner has, and use the negative feedback to correct the assumption. While these models are simplistic now, they might be a good first step towards a more complete model of human fairness.\nocite{honerkamp2021learning,mittal2019vision}

\section{Outlook}

As we expect robots to become ubiquitous agents in everyday human environments, we have to ensure that their interactions are physically sociable as well as fair towards the humans they interact with. In this paper, we have proposed to address these challenges not by attempting to learn positive behavior, but rather by learning to avoid negative behavior. We view this approach as more reliable than learning the positive, as it is hard to say what is perfectly sociable or fair, but it is easy to determine what is antisocial or unfair.
To move forward, we have provided references to existing works in safe human-robot interactions, as well as methods that can enable a robot to learn how \emph{not} to act.
With respect to developing \emph{fair} robots, we have exemplified how it will be important for a robot to recognize diverse groups and adapt its behavior to their needs. To aid fellow roboticists, we presented sociological and philosophical references which study the matter. From these sources we also find that our proposed inversion is something very human.

\footnotesize
\balance
\bibliographystyle{IEEEtran}
\bibliography{references.bib}

\begin{thebibliography}{10}
\providecommand{\url}[1]{#1}
\csname url@samestyle\endcsname
\providecommand{\newblock}{\relax}
\providecommand{\bibinfo}[2]{#2}
\providecommand{\BIBentrySTDinterwordspacing}{\spaceskip=0pt\relax}
\providecommand{\BIBentryALTinterwordstretchfactor}{4}
\providecommand{\BIBentryALTinterwordspacing}{\spaceskip=\fontdimen2\font plus
\BIBentryALTinterwordstretchfactor\fontdimen3\font minus
  \fontdimen4\font\relax}
\providecommand{\BIBforeignlanguage}[2]{{%
\expandafter\ifx\csname l@#1\endcsname\relax
\typeout{** WARNING: IEEEtran.bst: No hyphenation pattern has been}%
\typeout{** loaded for the language `#1'. Using the pattern for}%
\typeout{** the default language instead.}%
\else
\language=\csname l@#1\endcsname
\fi
#2}}
\providecommand{\BIBdecl}{\relax}
\BIBdecl

\bibitem{nelson2019bias}
G.~S. Nelson, ``Bias in artificial intelligence,'' \emph{North Carolina medical
  journal}, vol.~80, no.~4, pp. 220--222, 2019.

\bibitem{yu2019framing}
K.-H. Yu and I.~S. Kohane, ``Framing the challenges of artificial intelligence
  in medicine,'' \emph{BMJ quality \& safety}, vol.~28, no.~3, pp. 238--241,
  2019.

\bibitem{corradi2016determines}
C.~Corradi-Dell'Acqua, L.~Koban, S.~Leiberg, and P.~Vuilleumier, ``What
  determines social behavior? investigating the role of emotions, self-centered
  motives, and social norms,'' \emph{Front in human neuroscience}, 2016.

\bibitem{findlay2006links}
L.~C. Findlay, A.~Girardi, and R.~J. Coplan, ``Links between empathy, social
  behavior, and social understanding in early childhood,'' \emph{Early
  Childhood Research Quarterly}, vol.~21, no.~3, pp. 347--359, 2006.

\bibitem{ewer1968amicable}
R.~Ewer, ``Amicable behaviour,'' in \emph{Ethology of Mammals}, 1968.

\bibitem{leite2013social}
I.~Leite, C.~Martinho, and A.~Paiva, ``Social robots for long-term interaction:
  a survey,'' \emph{International Journal of Social Robotics}, 2013.

\bibitem{bicchi2015social}
A.~Bicchi and G.~Tamburrini, ``Social robotics and societies of robots,''
  \emph{The Information Society}, vol.~31, no.~3, pp. 237--243, 2015.

\bibitem{robert2018personality}
L.~Robert, ``Personality in the human robot interaction literature: A review
  and brief critique,'' in \emph{Proceedings of the 24th Americas Conference on
  Information Systems}, 2018, pp. 16--18.

\bibitem{lasota2017survey}
P.~A. Lasota, T.~Fong, J.~A. Shah \emph{et~al.}, \emph{A survey of methods for
  safe human-robot interaction}.\hskip 1em plus 0.5em minus 0.4em\relax Now
  Publishers, 2017.

\bibitem{charalampous2017recent}
K.~Charalampous, I.~Kostavelis, and A.~Gasteratos, ``Recent trends in social
  aware robot navigation: A survey,'' \emph{Robotics and Autonomous Systems},
  vol.~93, pp. 85--104, 2017.

\bibitem{mainprice2013human}
J.~Mainprice and D.~Berenson, ``Human-robot collaborative manipulation planning
  using early prediction of human motion,'' in \emph{IEEE/RSJ International
  Conference on Intelligent Robots and Systems}, 2013.

\bibitem{kirby2010social}
R.~Kirby, \emph{Social robot navigation}.\hskip 1em plus 0.5em minus
  0.4em\relax Carnegie Mellon University, 2010.

\bibitem{rios2015proxemics}
J.~Rios-Martinez, A.~Spalanzani, and C.~Laugier, ``From proxemics theory to
  socially-aware navigation: A survey,'' \emph{International Journal of Social
  Robotics}, vol.~7, no.~2, pp. 137--153, 2015.

\bibitem{bajcsy2017learning}
A.~Bajcsy, D.~P. Losey, M.~K. O’Malley, and A.~D. Dragan, ``Learning robot
  objectives from physical human interaction,'' 2017, pp. 217--226.

\bibitem{chisari2021correct}
E.~Chisari, T.~Welschehold, J.~Boedecker, W.~Burgard, and A.~Valada, ``Correct
  me if i am wrong: Interactive learning for robotic manipulation,'' \emph{IEEE
  Robotics and Automation Letters (RA-L)}, 2022.

\bibitem{caton2020fairness}
S.~Caton and C.~Haas, ``Fairness in machine learning: {A} survey,'' \emph{arXiv
  preprint arXiv:2010.04053}, 2020.

\bibitem{hamilton2019sexist}
M.~Hamilton, ``The sexist algorithm,'' \emph{Behavioral sciences \& the law},
  vol.~37, no.~2, pp. 145--157, 2019.

\bibitem{rudin2018age}
C.~Rudin, C.~Wang, and B.~Coker, ``The age of secrecy and unfairness in
  recidivism prediction,'' \emph{arXiv preprint arXiv:1811.00731}, 2018.

\bibitem{hamilton2019biased}
M.~Hamilton, ``The biased algorithm: Evidence of disparate impact on
  hispanics,'' \emph{Am. Crim. L. Rev.}, vol.~56, p. 1553, 2019.

\bibitem{bratman1992shared}
M.~E. Bratman, ``Shared cooperative activity,'' \emph{The philosophical
  review}, vol. 101, no.~2, pp. 327--341, 1992.

\bibitem{fu2021fair}
R.~Fu, M.~Aseri, P.~Singh, and K.~Srinivasan, ``“un” fair machine learning
  algorithms,'' \emph{Management Science}, 2021.

\bibitem{johnson2021algorithmic}
G.~M. Johnson, ``Algorithmic bias: on the implicit biases of social
  technology,'' \emph{Synthese}, vol. 198, no.~10, pp. 9941--9961, 2021.

\bibitem{hurtado2021learning}
J.~V. Hurtado, L.~Londo{\~n}o, and A.~Valada, ``From learning to relearning: A
  framework for diminishing bias in social robot navigation,'' \emph{Frontiers
  in Robotics and AI}, vol.~8, p.~69, 2021.

\bibitem{corbett2018measure}
S.~Corbett-Davies and S.~Goel, ``The measure and mismeasure of fairness: A
  critical review of fair machine learning,'' \emph{arXiv preprint
  arXiv:1808.00023}, 2018.

\bibitem{chang2020defining}
M.~L. Chang, Z.~Pope, E.~S. Short, and A.~L. Thomaz, ``Defining fairness in
  human-robot teams,'' in \emph{IEEE International Conference on Robot and
  Human Interactive Communication}, 2020, pp. 1251--1258.

\bibitem{lupfer2000folk}
M.~B. Lupfer, K.~P. Weeks, K.~A. Doan, and D.~A. Houston, ``Folk conceptions of
  fairness and unfairness,'' \emph{European Journal of Social Psychology},
  vol.~30, no.~3, pp. 405--428, 2000.

\bibitem{herrine2021folklore}
L.~Herrine, ``The folklore of unfairness,'' \emph{NYUL Rev.}, vol.~96, 2021.

\bibitem{ryan2006fairness}
A.~Ryan, ``Fairness and philosophy,'' \emph{social research}, pp. 597--606,
  2006.

\bibitem{finkel2001not}
N.~J. Finkel, \emph{Not fair!: The typology of commonsense unfairness.}\hskip
  1em plus 0.5em minus 0.4em\relax American Psychological Association, 2001.

\bibitem{mikula1990people}
G.~Mikula, B.~Petri, and N.~Tanzer, ``What people regard as unjust: Types and
  structures of everyday experiences of injustice,'' \emph{European journal of
  social psychology}, vol.~20, no.~2, pp. 133--149, 1990.

\bibitem{zhang2021multi}
K.~Zhang, Z.~Yang, and T.~Ba{\c{s}}ar, ``Multi-agent reinforcement learning: A
  selective overview of theories and algorithms,'' \emph{Handbook of
  Reinforcement Learning and Control}, pp. 321--384, 2021.

\bibitem{tejwani2022social}
R.~Tejwani, Y.-L. Kuo, T.~Shu, B.~Katz, and A.~Barbu, ``Social interactions as
  recursive mdps,'' in \emph{Proceedings of the 5th Conference on Robot
  Learning}, vol. 164, 2022, pp. 949--958.

\bibitem{honerkamp2021learning}
D.~Honerkamp, T.~Welschehold, and A.~Valada, ``Learning kinematic feasibility
  for mobile manipulation through deep reinforcement learning,'' \emph{IEEE
  Robotics and Automation Letters}, 2021.

\bibitem{mittal2019vision}
M.~Mittal, R.~Mohan, W.~Burgard, and A.~Valada, ``Vision-based autonomous uav
  navigation and landing for urban search and rescue,'' \emph{arXiv preprint
  arXiv:1906.01304}, 2019.

\end{thebibliography}

\end{document}